# Adaptive Wind Driven Optimization Trained Artificial Neural Networks

Zikri Bayraktar
Schlumberger-Doll Research Center
Cambridge, MA USA
zikribayraktar@ieee.org

*Abstract*—This paper presents the application of a newly developed nature-inspired metaheuristic optimization method, namely the Adaptive Wind Driven Optimization (AWDO), to the training of feedforward artificial neural networks (NN) and presents a discussion into the future research of AWDO implementation in Deep Learning (DL). Application example of digit classification with MNIST dataset reveals interesting behavior of the derivative-free AWDO method compared to steepest descent method where results and future work on the implementation of AWDO in deep neural networks are discussed.

## I. INTRODUCTION

The artificial neural networks are one of the many methods in machine learning [1, 2] and they are becoming increasingly preferred method thanks to the success of recent developments in parallel computing architectures and GPUs which allows deeper neural networks to be utilizes with relative ease. This progress opened up the field of deep learning [3] for significant growth but deep neural networks also come with a new set of challenges such as vanishing or exploding gradients in the computation of the weights. In this work, the weights of NNs are optimized via a newly developed metaheuristic optimization method named the Adaptive Wind Driven Optimization [4], which is a gradient-free optimization algorithm eliminating the vanishing or exploding gradient problems completely.

In the next section, a brief overview of the AWDO is provided followed by the details of the test example on MNIST handwritten digit recognition. Finally, results comparing the steepest descent optimized NN and AWDO optimized NN is presented along with a discussion on the future research in application of the AWDO to deep learning neural networks.

## II. ADAPTIVE WIND DRIVEN OPTIMIZATION

The Wind Driven Optimization (WDO) method is a nature-inspired gradient-free metaheuristic optimization method, which was introduced in [5]. It is a population based iterative algorithm based on the abstraction of the atmospheric dynamics equations explaining the wind motion in hydrostatic balance. This motion can be represented by the Eulerian description of a point air parcel obeying the Newton's second law of motion. Based on this abstraction, the position and velocity of the air parcel can be

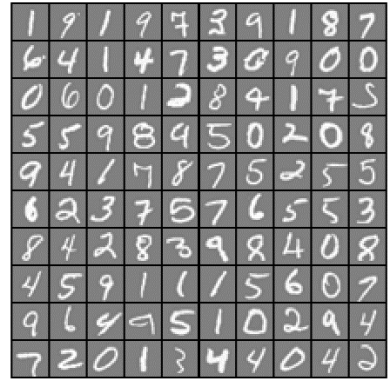

Fig. 1. A sample subset from the MNIST dataset.

computed by the update equations at each iteration as follows:

$$\vec{u}_{new} = (1-\alpha)\vec{u}_{cur} - g(\vec{x}_{cur}) + \left|1 - \frac{1}{i}\right| R\,T\,(x_{max} - x_{cur}) + \frac{c\,u_{cur}^{otherdim}}{i}$$

$$\vec{x}_{new} = \vec{x}_{cur} + (\vec{u}_{new} \times \Delta t)$$

where $i$ represents the rank of the air parcel among population members, $\alpha$ is the friction coefficient, $g$ is gravitational constant, $R$ is universal gas constant, $T$ is temperature and $c$ represents rotation of the Earth. The updated velocity ($u_{new}$) is computed at each dimension and is affected by current velocity ($u_{cur}$), current location in the search domain ($x_{cur}$), distance from the best pressure point that found in the previous iterations ($x_{max}$) as well as a velocity at another dimension ($u_{cur}^{otherdim}$). In WDO, pressure is analogous to the cost/loss in general, which is used to determine the success of air parcels at meeting the optimization goals. Similar to cost, low pressure is a good solution and high pressure is a bad solution. The position of each air parcel is limited to [-1, 1] and the parcel velocity is limited to $V_{max} = \pm |0.3|$. After the velocity ($u_{new}$) is updated, the position ($x_{new}$) is updated where a unity time step, $\Delta t = 1$, is assumed. The velocity and the position of each parcel are iteratively updated at each iteration based on the pressure value and they move from high pressure regions to lower pressure regions similar to the wind in out atmosphere. Termination can be achieved by setting a maximum number of iterations or a predefined pressure value target. The classical WDO algorithm requires fine tuning of four inherent parameters in the update equations. To alleviate the problems with tuning the parameters, the Adaptive WDO is introduced and demonstrated to have superior performance compared to classical WDO [4]. In AWDO, we combined the classical WDO with a black box solver, namely Covariance Matrix Adaptation Evolutionary Strategy (CMAES), to tune the inherent parameters of the WDO so that the inherent parameters are adaptively selected at each iteration. This created an adaptive algorithm suitable for many types of optimization problems.

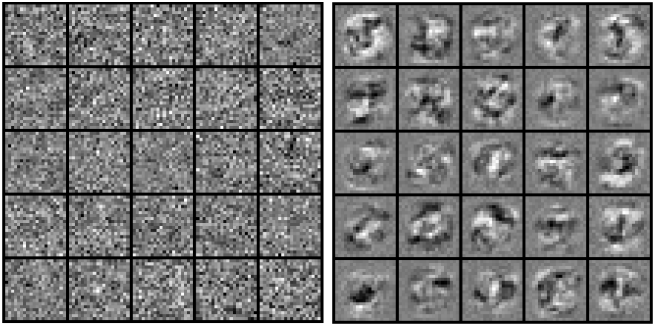
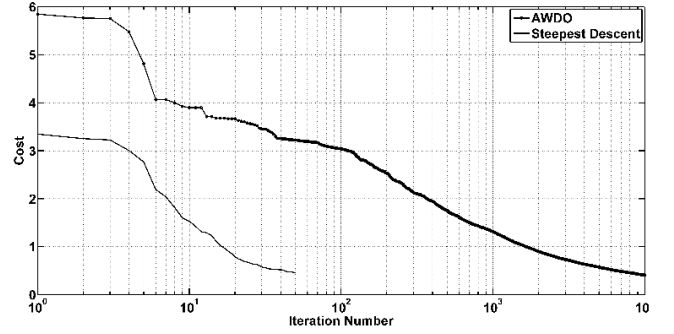

Fig. 2. Visualization of the weights in the hidden layer with 25 neurons. Weights of the AWDO optimized NN shown on the left and gradient descent optimized NN on the right.

Fig. 3. Training set cost (pressure) function comparison; AWDO performance vs steepest descent performance over the number of iterations.

## III. AWDO TRAINED NN FOR MNIST DIGIT RECOGNITION

One of the well-known benchmarks in the design of neural networks is the MNIST handwritten digit recognition datasets. In this section, I compare the results from a three layer feedforward NN trained by the AWDO against the same network model trained by the backpropagation using gradient descent (GD). The input data set contains 5000 training examples of 20 pixel by 20 pixel grayscale images of handwritten digits as shown in Fig. 1. This input creates an input layer of 400 neurons. The hidden layer consists of 25 neurons and the output layer is 10 neurons to output one-hot encoding of the 10 digits. In this example, sigmoid activation function and a regularized cost function, $J(\theta)$, is utilized as shown below:

$$J(\theta) = \frac{1}{m}\sum_{i=1}^{m}\sum_{k=1}^{K}[-y_k^{(i)}\log((h_\theta(x^i))_k) - (1-y_k^{(i)})\log((h_\theta(x^i))_k)] + \frac{\lambda}{2m}\left[\sum_{j=1}^{25}\sum_{k=1}^{400}(\theta_{j,k}^{(1)})^2 + \sum_{j=1}^{10}\sum_{k=1}^{25}(\theta_{j,k}^{(2)})^2\right]$$

where $\theta$ represents the weights, $h_\theta(x^{(i)})_k$ is the sigmoid activation function, $\lambda$ is the regularization constant of 0.01, $x^i$ is the input to the activation function, $y_k$ is the label data set and $m$ is the number of training data. The initial values for the weights and biases are randomly picked from a uniform distribution in the range of [-0.12, 0.12].

The AWDO optimized NN achieves 95% accuracy after 10,000 iterations using a population of 25 air parcels, whereas GD achieves the same 95% digit recognition accuracy in the training set after 50 iterations as seen in Fig. 3. This constitutes a 2-orders of magnitude convergence speed between the two algorithms. Fig. 2 illustrates the 25 neurons in the hidden layer, as we can see the multi-pole nature of the solution of this problems by the widely different weights configuration illustrated in Fig. 2.

## IV. DISCUSSION AND FUTURE WORK

In this work, the AWDO algorithm is applied to the optimization of the weights and biases of a three layer artificial neural network for the MNIST digit recognition. Comparison of the weights on the hidden layer in Fig. 2 demonstrates the multi-pole nature of the solutions and convergence speed comparison results shown in Fig 3 illustrates that for training of a shallow network, gradient descent method could converge to a solution faster; two-orders or magnitude faster to be exact.

As future work, we are comparing the tolerance of the both networks to the noisy data. In addition, we are applying the AWDO based optimization to reinforcement learning (RL) per the success of neuroevolution in this field [7]. OpenAI Gym platform provides RL benchmarks [8], like Atari Games, and we will be presenting the results of this work. Compared to evolutionary algorithms, AWDO has a cooperating population instead of competing, which can provide additional benefits through information sharing among the population members.